\pdfoutput=1

\documentclass[11pt]{article}

\usepackage[]{formats/eacl2023}

\usepackage{times}
\usepackage{latexsym}
\usepackage{amsmath, amssymb, amsfonts, gensymb, bm}
\usepackage{centernot}
\usepackage{enumerate}
\usepackage{enumitem}
\usepackage{graphicx}
\usepackage{subcaption}
\usepackage{multirow}
\usepackage{array}
\usepackage{arydshln}

\usepackage{makecell}
\usepackage{soul}
\usepackage{xspace}
\usepackage{comment}

\usepackage[T1]{fontenc}

\usepackage[utf8]{inputenc}

\usepackage{microtype}

\newcommand{\astar}{$^{\star}$\xspace}

\newcommand{\textev}[1]{\textsc{#1}}

\newcommand{\excep}{\textev{exception}\xspace}
\newcommand{\inst}{\textev{instantiation}\xspace}
\newcommand{\Excep}{\textev{Exception}\xspace}
\newcommand{\Inst}{\textev{Instantiation}\xspace}
\newcommand{\exceps}{\textev{exceptions}\xspace}
\newcommand{\insts}{\textev{instantiations}\xspace}
\newcommand{\Exceps}{\textev{Exceptions}\xspace}
\newcommand{\Insts}{\textev{Instantiations}\xspace}
\newcommand{\Eshort}{\textev{Excep.}\xspace}
\newcommand{\Ishort}{\textev{Inst.}\xspace}
\newcommand{\evtxt}{exemplars\xspace}
\newcommand{\ev}{\textev{exemplars}\xspace}
\newcommand{\Ev}{\textev{Exemplars}\xspace}
\newcommand{\evsing}{\textev{exemplar}\xspace}
\newcommand{\pragneg}{exoproperty\xspace}
\newcommand{\pragnegs}{exoproperties\xspace}

\usepackage{amsthm}
\newtheorem*{remark}{Definition}

\title{Penguins Don't Fly:\\Reasoning about Generics through Instantiations and Exceptions}

\author{Emily Allaway$^{\star}$ \quad Jena D. Hwang$^{\dagger}$ \quad Chandra Bhagavatula$^{\dagger}$\\ {\bf Kathleen McKeown$^{\star}$ \quad Doug Downey$^{\dagger\mathparagraph}$ \quad Yejin Choi$^{\dagger\ddagger}$}\\
  $^{\star}$Columbia University, USA \\
  $^{\dagger}$Allen Institute for Artificial Intelligence, USA\\
  $^{\mathparagraph}$Northwestern University, USA\\
  $^{\ddagger}$Paul G. Allen School of Computer Science \& Engineering, University of Washington, USA\\
  \texttt{eallaway@cs.columbia.edu}}

\begin{document}
\maketitle
\begin{abstract}
Generics express generalizations about the world (e.g., birds can fly)
that are not universally true (e.g., newborn birds and penguins cannot fly). Commonsense knowledge bases, used extensively in NLP, encode some generic knowledge but rarely enumerate such exceptions and
knowing when a generic statement holds or does not hold true is crucial for developing a comprehensive understanding of generics.
We present a novel framework informed by linguistic theory to generate \ev---specific cases when a generic holds true or false.
We generate ${\sim}19k$ \evtxt for ${\sim}650$ generics and show that our framework outperforms a strong GPT-3 baseline by $12.8$ precision points.
Our analysis highlights the importance of
linguistic theory-based controllability for generating \evtxt, the insufficiency of knowledge bases as a source of \evtxt, and the challenges \evtxt pose for the
task of natural language inference.

\end{abstract}

\section{Introduction}
Generics express generalizations (e.g., birds can fly) that allow humans to reason and act with incomplete world knowledge~\citep{asher1995some}. 
Generics allow us 
to draw plausible inferences about individuals (e.g., Polly is a bird, so Polly can fly) \textit{even when we know of counterexamples} (e.g., penguins cannot fly). 
Despite the utility of generalizations, knowledge of counterexamples is necessary for modeling generics and effectively reasoning with them in computational systems. 

Recent studies of generics~\citep[e.g.,][]{Bhagavatula2022I2D2IK,Bhakthavatsalam2020GenericsKBAK} and commonsense KBs~\citep[e.g.,][]{speer2017conceptnet} provide repositories of generic knowledge. However, these resources rarely mention \exceps (i.e., counterexamples) or \insts (i.e., cases where the generic holds); collectively \ev. For systems using these resources as a source of world knowledge, such incomplete information can lead to incorrect deductions (e.g., if Polly is a penguin, which is a bird, then inferring that Polly can fly because birds can fly is false\footnote{penguin is a bird $\wedge$ birds can fly $\implies$ penguins can fly}).
Therefore, we propose a novel computational framework that operationalizes linguistic theories in order to automatically 
generate \ev.

\begin{figure}
    \centering
    \includegraphics[width=0.95\columnwidth]{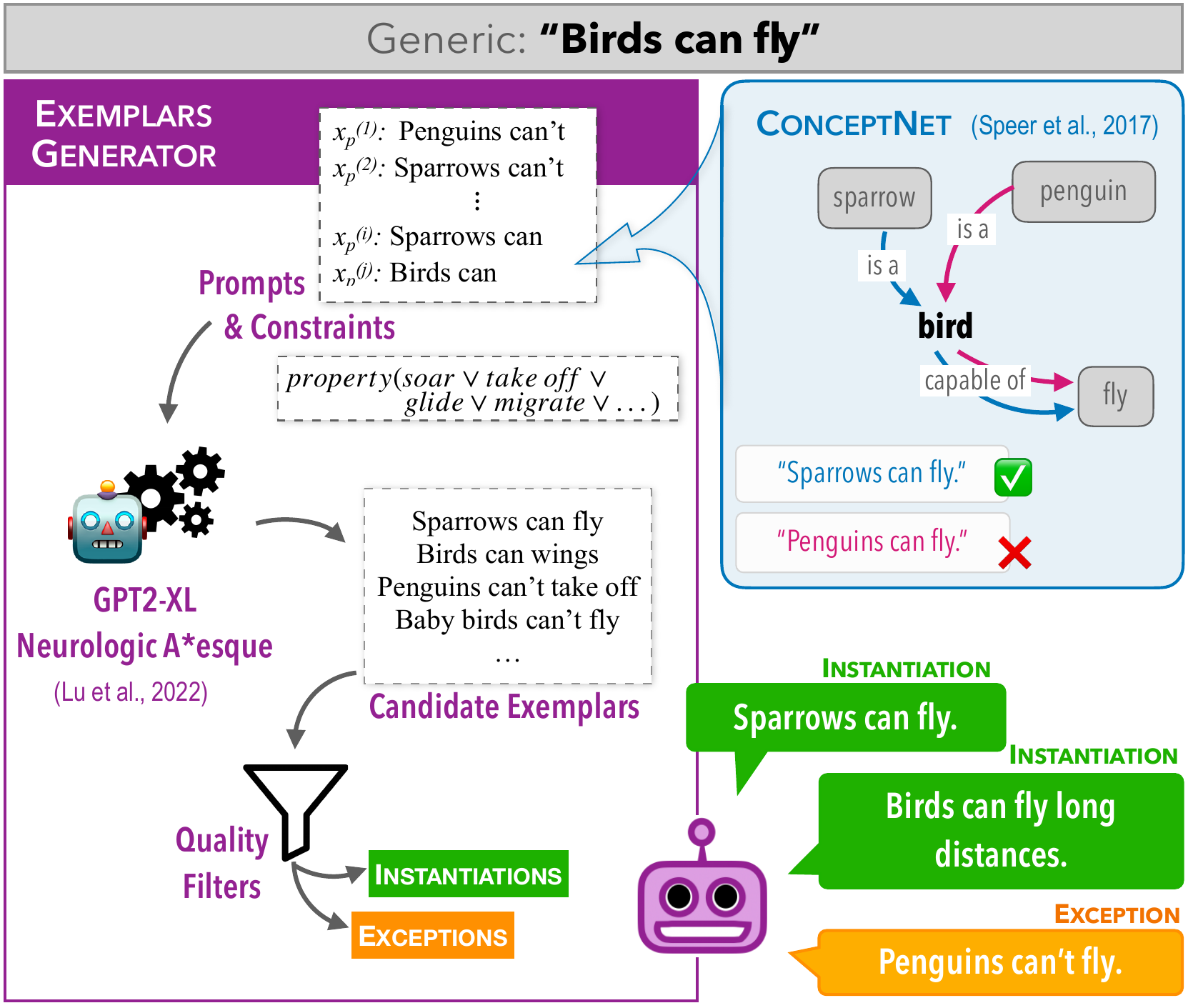}
    \caption{
    We present \textbf{\ev generator}: 
    given a generic like ``Birds can fly'' it generates truthful statements where the generic does (\insts) and does not (\exceps) hold. 
    We extract commonsense knowledge (e.g., from ConceptNet \citep{speer2017conceptnet}) 
    in
    linguistically-informed prompts and constraints for constrained generation \citep{lu2021neurologic}. We use trained discriminators 
    to filter for quality.
    }
    \label{fig:intro}
\end{figure}

In our work, we unify two distinct linguistic theories and use linguistic theory-guided decoding to generate \ev.
Although large-scale neural language models  such as GPT-3~\citep{brown2020language} have been increasingly successful in few-shot text generation tasks, such generation is both expensive and not easily controllable. Therefore, we instead use 
the constrained generation algorithm Neurologic A{\astar}esque~\cite{lu2021neurologic}, which can be applied to any auto-regressive language model; we choose to use GPT-2~\cite{radford2019language}.
In this manner, we generate $12562$ \insts and $6297$ \exceps for $\sim$650 generic statements from~\citet{Bhagavatula2022I2D2IK}. 
We conduct human evaluation of the generations and show that our system outperforms few-shot generation by GPT-3 by $12.8$ precision points. Our analyses demonstrate not only the importance of linguistic modeling for generating \ev and the insufficiency of KBs as a source of \ev, but also the challenges \ev pose for natural language reasoning.

\begin{table*}[t]
    \centering
    \scalebox{0.77}{
    \setlength\tabcolsep{2pt}
    \begin{tabular}{lllll}
    \hline
        & \textbf{Category} & \textbf{Generic ($G$)}  & \textbf{\inst} & \textbf{\excep} \\
    \hline
    
    \textbf{(a)}& \textbf{quasi-def} & ``{\color{blue}Stars} produce {\color{magenta}radiation}'' &  ``The sun produces radiation'' & ``Stars produce light''\\
    \cdashline{3-5}
        & &\ {\color{blue}$K(x)$} $\wedge r(x,y) \implies$ {\color{magenta}$P(y)$} & 
                \ $K(x) \wedge r(x,y) \wedge P(y)$ &  
                \ $K(x) \wedge r(x,y) \wedge {\nsim}P(y)$\\

    \hline

        & & ``{\color{blue}Birds} can {\color{magenta}fly}'' & ``Owls can fly'' &  ``Penguins can't fly''\\ 
        \textbf{(b)} & \textbf{principled}  & ``{\color{blue}Sharks} attack {\color{magenta}swimmers}''  & ``Threatened sharks attack swimmers'' & \makecell[l]{``Sharks don't attack  
        swimmers\\ \quad in the shallows''}\\ 

        \cdashline{3-5}
        & & \ {\color{blue}$K(x)$} $\wedge$ {\color{magenta}$P(y)$} $\implies r(x, y)$ & 
        \ $K(x) \wedge r(x,y) \wedge P(y)$ &  
        \ $K(x) \wedge \neg r(x,y) \wedge P(y)$\\         

    \hline
    
        & & & & ``Cars have CD Players'' \\

     \cdashline{5-5}
       \textbf{(c)} & \makecell[l]{\textbf{characterizing}} & \multirow{2}{*}{
         \begin{tabular}{p{45mm}}
                ``{\color{blue}Cars} have {\color{magenta}radios}''  \\ 
                \cdashline{1-1}
                \makecell[l]{$L_G$ is ambiguous} \\
            \end{tabular}   
        } & \multirow{2}{*}{
        \begin{tabular}{l}
                ``2014 Prius model C has a radio ''  \\ 
                \cdashline{1-1}
                \ $K(x) \wedge r(x,y) \wedge P(y)$   \\
            \end{tabular}
        } &
                \ $K(x) \wedge  r(x,y) \wedge {\nsim}P(y)$\\         

        & & & & ``Newer cars don't have radios'' \\  

     \cdashline{5-5}
       & & & &
                \ $K(x) \wedge  \neg r(x,y) \wedge P(y)$\\

    \hline
    \end{tabular}
    }
    \caption{
    We define three categories of generics with their \ev.
    The logical forms for the generic ($L_G$) and its 
    \ev
    are also below the examples. 
    Using these categories we formulate templates for generating \ev (see Table \ref{tab:patts}). 
    $K$ is the concept ({\color{blue}blue}), $P$ the property ({\color{magenta}pink}). See \S\ref{sec:evlogic} for \pragneg ${\nsim}P$.
    }
    \label{tab:excepttypes}
\end{table*}

Our contributions are as follows: \textbf{(1)} we present a novel framework grounded in linguistic theory for representing generics and their \ev, \textbf{(2)} we present the first method to automatically generate generic \ev and show it outperforms a competitive baseline based on GPT-3 
and \textbf{(3)} we present analysis 
showing the importance of explicit linguistic modeling for this task and 
the insufficiency of current NLI methods for
generics.
Our system and data are publicly available\footnote{\url{https://github.com/emilyallaway/generics-exemplars}}.

\section{Related Work}
\label{sec:relwork}

\paragraph{Theory} Generics have been studied extensively in semantics, philosophy, and psychology  to develop a single logical form for all generics~\citep{lewiskeenan1975,Carlson1977ReferenceTK,Carlson1989OnTS,krifka1987outline} or a probabilistic definition~\citep{cohen1996think,cohen1999generics,cohen2004generics,kochari2020generics}, 
categorize generics~\citep{leslie2007generics,leslie2008generics,khemlani2009generics}, and analyze specific types~\citep{prasada2006principled,prasada2009representation,haward2018development,mari2012genericity,krifka2012definitional}. Mechanisms to tolerate \exceps have also been proposed~\citep{kadmon1993any,greenberg2007exceptions,lazaridou2013genericity} but these are primarily theoretical and use carefully chosen examples. In contrast, our work combines
these
\excep tolerance mechanisms with generic categorization and proposes a novel, large-scale, computational framework for  \ev.

\paragraph{Commonsense Knowledge} While large-scale CKBs capture a range of commonsense knowledge~\citep{speer2017conceptnet,Sap2019ATOMICAA,Forbes2020SocialC1,Hwang2021COMETATOMIC2O}, they contain necessarily incomplete (i.e., the open-world assumption~\citep{Reiter1978OCW}) \textit{general} knowledge.
Furthermore, although recent works have created KBs specifically of generics~\cite{Bhakthavatsalam2020GenericsKBAK,Bhagavatula2022I2D2IK} and proposed methods to identify generics in text~\cite{Friedrich2015AnnotatingGA,Friedrich2016SituationET}, they do not identify or model \ev.
In our work, 
we focus directly on automatically generating \ev, providing richer commonsense knowledge.

The application 
of generics to specific 
individuals
is influenced by prototypicality~\citep{Rips1975InductiveJA,Osherson1990CategoryBasedI}, with small sets of prototypical norms collected in cognitive science for a range of kinds ~\citep{Devereux2014TheCF,McRae2005SemanticFP,Overschelde2004CategoryNA}. However, recent work has shown that neural models have only moderate success at mimicking human prototypicality~\citep{Misra2021DoLM,Boratko2020ProtoQAAQ} or producing commonsense facts without guidance~\citep{Petroni2019LanguageMA}.
Hence,
we combine
neural models
with a KB of concepts,
using linguistic-theory-guided decoding,
to generate generics \ev.

\paragraph{Reasoning} Reasoning with generics is
closely related to non-monotonic reasoning~\citep{Ginsberg1987ReadingsIN,ginsberg1987}; specifically default inheritance reasoning~\citep{Brewka1987TheLO,Hanks1986DefaultRN,Horty1988MixingSA,Imielinski1985ResultsOT,Poole1988ALF,reiter1978,Reiter1980ALF}. 
Contrary to the proposed solutions for linguistic tests
on default inheritance reasoning \citep[e.g.,can a conclusion about inheritance be inferred based on provided evidence?]{lifschitz1989}, later works showed that the presence of generics \ev in the evidence impacts what humans perceive as the correct answer~\citep{elio1996reasoning,pelletier2005case,pelletier2009all}.
These results 
highlight the importance of identifying generics and 
accurately modeling their relationships in machine reasoning. 

While natural language inference (NLI), a form of deductive reasoning
well-studied in NLP (i.a.,~\citet{dagan2013recognizing,bowman2015large}), captures notions of inference, studies on non-monotonic reasoning and NLI are limited~\citep{wang2018glue,cooper1994fracas,Yanaka2019CanNN,Yanaka2019HELPAD,rudinger2020thinking} and do not include default inheritance reasoning. Therefore, in this work we analyze the interactions between generics \ev and NLI and highlight the importance of modeling this relationship in machine reasoning.

\section{Framework for \Ev}
\label{sec:defs}

A generic statement describes a \textit{relation} between a \textit{concept}  and a \textit{property}.
Usually, a \textbf{concept} $\bm{K}$ is a type or kind (e.g., bird) while a \textbf{property} $\bm{P}$ is an ability (e.g., fly) or quality (e.g., feathered).
Note that statements containing explicit quantification (e.g., ``\textit{Most} birds can fly'') are generally \textit{not} considered generics~\citep{Carlson1977ReferenceTK,krifka1995genericity} and are therefore excluded from this study (see \S\ref{sec:genquant} for further discussion).

In our framework, we first 
categorize (\S\ref{sec:gencat}) generics and derive their logical forms (\S\ref{sec:genlogic}). These logical forms for generics serve as our basis for formulating \ev (\S\ref{sec:evlogic}) and designing templates suitable for generation (\S\ref{sec:tempder}).

\subsection{Generic Category Definitions}\label{sec:gencat}
We categorize a generic based on the type of property it describes. In particular, by unifying theories from linguistics and philosophy\footnote{Description of the theories is provided in \S\ref{app:gendef}}, we split generics into three categories (see examples in Table~\ref{tab:excepttypes}). A generic has a particular category if:
\begin{enumerate} 
\setlength{\itemindent}{0em}
\itemsep 0em
\vspace{-5pt}
    \item[(a)] \textbf{Quasi-definitional}: the property is essential to a concept~\citep{khemlani2009generics}.
    \item[(b)] \textbf{Principled}: the property has a strong association with the concept. This includes both properties with a principled association to a concept (e.g., flying is viewed as inherent to birds, although it is not essential in reality)~\citep{prasada2006principled,prasada2009representation,haward2018development} and properties that are uncommon and often dangerous~\citep{leslie2017original}.
    \item[(c)] \textbf{Characterizing (char.)}: there is only a non-accidental relationship between the property and concept (e.g., based only on absolute or relative prevalence among concepts) \citep{leslie2007generics,leslie2008generics}. 
\end{enumerate}

\subsection{Logical Forms for Generics}\label{sec:genlogic}
We propose logical forms $L_G$ that are used to represent an individual generic $G$. \textbf{Each generic category has a distinct logical form} (see Table~\ref{tab:excepttypes}). 
For quasi-definitional generics, since the property is defining we assert that the property is logically implied by 
the concept and relationship together (i.e., $K \wedge r \implies P$). In contrast, for principled generics
we assert that the concept and property together 
logically imply the relationship (i.e., $K \wedge P \implies r$). Finally, for characterizing generics, the logical form depends on 
whether the generic is interpreted as quasi-definitional or principled.
Logical forms with examples are shown in Table~\ref{tab:excepttypes}. 
Note that in a logical $L_G$, the concept $K$ (property $P$) is satisfied by \textit{both an individual or subtype} of $K$ ($P$)\footnote{For example, \textsc{Bird}($x$) is true for $x_1=$``my parrot'', $x_2=$``owls'', and $x_3=$``these birds'' since all \textit{are} birds.}.

\subsection{Constructing \Ev}\label{sec:evlogic}
We now define generics \ev, deriving them from the logical form of a generic.

\paragraph{\Insts}
For a generic, \insts are contextually relevant \textbf{members of the concept 
with
the desired property}. Formally,
\begin{remark}[\Insts]
An \inst 
satisfies $L_G'$ (i.e., $L_G$ with implication replaced by conjunction\footnote{Replacing the implication with a conjunction excludes instances which satisfy $L_G$ by satisfying \textit{only} the right side of the implication (e.g., $\textsc{Bird}(x) \wedge \textsc{Fly}(y) \implies \textit{can}(x,y)$ is logically true for  $(x,y)=($ airplanes, fly) but this is not a valid \inst).}).
\end{remark}
\noindent
\textbf{For example}, if we have 
\setlength{\abovedisplayskip}{2pt}
\setlength{\belowdisplayskip}{2pt}
\begin{align*}
L_G\text{:} \ & \textsc{Bird}(x) \wedge \textsc{Fly}(y) \implies \textit{can}(x, y)\\
\text{so, } L_G'\text{:} \ & \textsc{Bird}(x) \wedge \textsc{Fly}(y) \wedge \textit{can}(x, y)
\end{align*}
\noindent then $(x,y)$ = (``owls'', ``fly'') satisfies $L_G'$. Note that \textbf{\insts all have the same logical form} regardless of the generic category (see Table~\ref{tab:excepttypes}).

\paragraph{\Exceps}
An \excep 
counters one of two interpretations of the generic (see \S\ref{appsec:evdef} for an in-depth discussion). 
\Exceps are members of the concept either \textbf{(i) \textit{without} the generic property}~\citep{greenberg2007exceptions} or (\textbf{ii) \textit{with} an alternative property} (i.e., \textbf{\pragneg}) when the generic property is essential to the concept. For example, ``penguins can't fly'' (for the generic ``birds can fly'') counters the interpretation that ``\textit{all} birds can fly''. In contrast, ``stars produce light'' (for the generic ``stars produce radiation'') counters the interpretation that ``stars produce \textit{only} radiation''. Note that in the latter case, there are no specific stars that do not produce radiation.
More formally,
\begin{remark}[\Exceps]
An \excep satisfies the logical form ${\nsim}L_G$ (i.e., $\neg L_G$ where $\neg P$ replaced by ${\nsim}P$ when applicable). We define $\bm{{\nsim}P}$ as \textbf{the \pragneg} of $P$: a property adjacent to $P$ that is \textbf{not} $P$ but \textbf{is} contextually relevant to $P$.
\end{remark}
\noindent
\textbf{For example},
if the logical form $L_G$ is
\begin{equation*}
\setlength{\abovedisplayskip}{3pt}
\setlength{\belowdisplayskip}{3pt} 
\textsc{Star}(x) \wedge \textit{produce}(x,y) \implies \textsc{Radiation}
\end{equation*}
then
${\nsim}L_G$ is
\begin{equation*}
\setlength{\abovedisplayskip}{3pt}
\setlength{\belowdisplayskip}{3pt} 
\textsc{Star}(x) \wedge \textit{produce}(x,y) \wedge {\nsim}\textsc{Radiation}(y)
\end{equation*}
and so $(x,y)$ = (``stars'', ``light'') satisfies ${\nsim}L_G$,
while $(x, y)$ = (``stars'', ``movies'') does not. Notice that the latter pair is invalid because ``movies'' is not a relevant alternative to ``radiation'' since it is not informative about the generic.

Since a generic's \exceps depend on its logical form $L_G$, they are also dependent on the generic's category (see Table~\ref{tab:excepttypes}).
In particular, the \exceps for quasi-defintional generics are individuals with alternative properties (since the property is viewed as essential) while for principled generics the \exceps are individuals without the generic property.  
\Exceps for characterizing generics are, like the logical forms, dependent on the generic's interpretation (see \S\ref{sec:genlogic})

\subsection{Logical Forms to Templates}
\label{sec:tempder}
Based on our proposed formulae (Table~\ref{tab:excepttypes}) for \ev 
we define seven templates for generation (Table~\ref{tab:patts}). Each template
expresses a set of instances
that satisfy the logical form of an \textev{exemplar}.
Each template has two sets of content specifications: for the \textit{input} and for the \textit{completion} (i.e., the decoder output).

For \textbf{\insts, we define three templates} with subtypes of the concept, property, or both. For \textbf{\exceps we have four templates}, using subtypes of either the concept \textit{or} property (but not both)
in order to avoid 
irrelevant or uninformative instances. For example, for the generic ``Birds can fly'', the sentence ``Penguins can't fly long distances'' (which has subtypes of both the concept and property) is \textit{uninformative} because it doesn't mean penguins can't fly \textit{in general} (e.g., 
they might still be able to fly short distances).

\begin{table*}[t]
\scalebox{0.9}{
\centering
\small
\begin{tabular}{ll|ll|lll|r}
    \hline
    & \textbf{Categories} & \multicolumn{2}{c|}{\textbf{Template}} & \multicolumn{3}{c|}{\textbf{Example}}\\
    & & \textit{input} & \textit{comp.} & Generic & Prompt $x_p$ & Constraints $\mathcal{C}$\\
    \hline
    \multirow{4}{*}{\Eshort} & 
    \multirow{2}{*}{
    \begin{tabular}[c]{@{}l@{}}quasi-def\\ \quad\& char\end{tabular}
    } & \texttt{{\color{blue}K} + $r$} & {\color{magenta}\texttt{${\nsim}$P}} &  \multirow{2}{*}{Stars produce radiation}& 
    ``{\color{blue}Stars} \textit{produce}'' & {\color{magenta}$\neg$radiation $\wedge \neg $ x-rays $\wedge$ ...} & 
    \texttt{t1}\\
    
    & & \texttt{{\color{blue}K$_{sub}$} + $r$} & {\color{magenta}\texttt{${\nsim}$P}} &  &
    ``{\color{blue}Sun}  \textit{produces}'' & {\color{magenta}$\neg$radiation $\wedge \neg $ x-rays $\wedge$ ...} & 
    \texttt{t2}\\
    
    \cdashline{2-8}
    & \multirow{2}{*}{
    \begin{tabular}[c]{@{}l@{}}principled\\\quad\& char\end{tabular}
    } & \texttt{{\color{blue}K} + $\neg r$} &   {\color{magenta}\texttt{P$_{sub}$}} & \multirow{2}{*}{Birds can fly} & 
    ``{\color{blue}Birds} \textit{can't}'' & {\color{magenta}migrate $\vee$ soar $\vee$ glide $\vee$...} & 
    \texttt{t3}\\ 
    
    & & \texttt{{\color{blue}K$_{sub}$} +  $\neg r$} & {\color{magenta}\texttt{P}} & &
    ``{\color{blue}Penguins} \textit{can't}'' & {\color{magenta}fly $\vee$ flying $\vee$...} & 
    \texttt{t4}\\

    \hline
    \multirow{3}{*}{\Ishort} 
    & \multirow{3}{*}{all} & \texttt{{\color{blue}K$_{sub}$} + $r$} & {\color{magenta}\texttt{P}} & \multirow{3}{*}{Birds can fly}& ``{\color{blue}Sparrows} \textit{can}'' & {\color{magenta}fly $\vee$ flying $\vee$...} & \texttt{t5} \\
    & & \texttt{{\color{blue}K} + $r$} & {\color{magenta}\texttt{P$_{sub}$}} & & ``{\color{blue}Birds} \textit{can}'' & {\color{magenta}migrate $\vee$ soar $\vee$ glide $\vee$...}& \texttt{t6}\\
    & & \texttt{{\color{blue}K$_{sub}$} + $r$} &   {\color{magenta}\texttt{P$_{sub}$}} & & ``{\color{blue}Sparrows} \textit{can}'' & {\color{magenta}migrate $\vee$ soar $\vee$ glide $\vee$...}& \texttt{t7}\\
    \hline
\end{tabular}
}
\caption{Templates for generating 
\ev,
derived from their logical forms (\S\ref{sec:evlogic}).
$sub$ indicates a subtype, \texttt{K} the concept, \texttt{P} the property, ${\nsim}$\texttt{P} its \pragneg
(\S\ref{sec:genlogic}). \textit{comp} is the completion.
}
\label{tab:patts}
\end{table*}

\begin{figure}[t]
    \centering
    \includegraphics[width=.85\columnwidth]{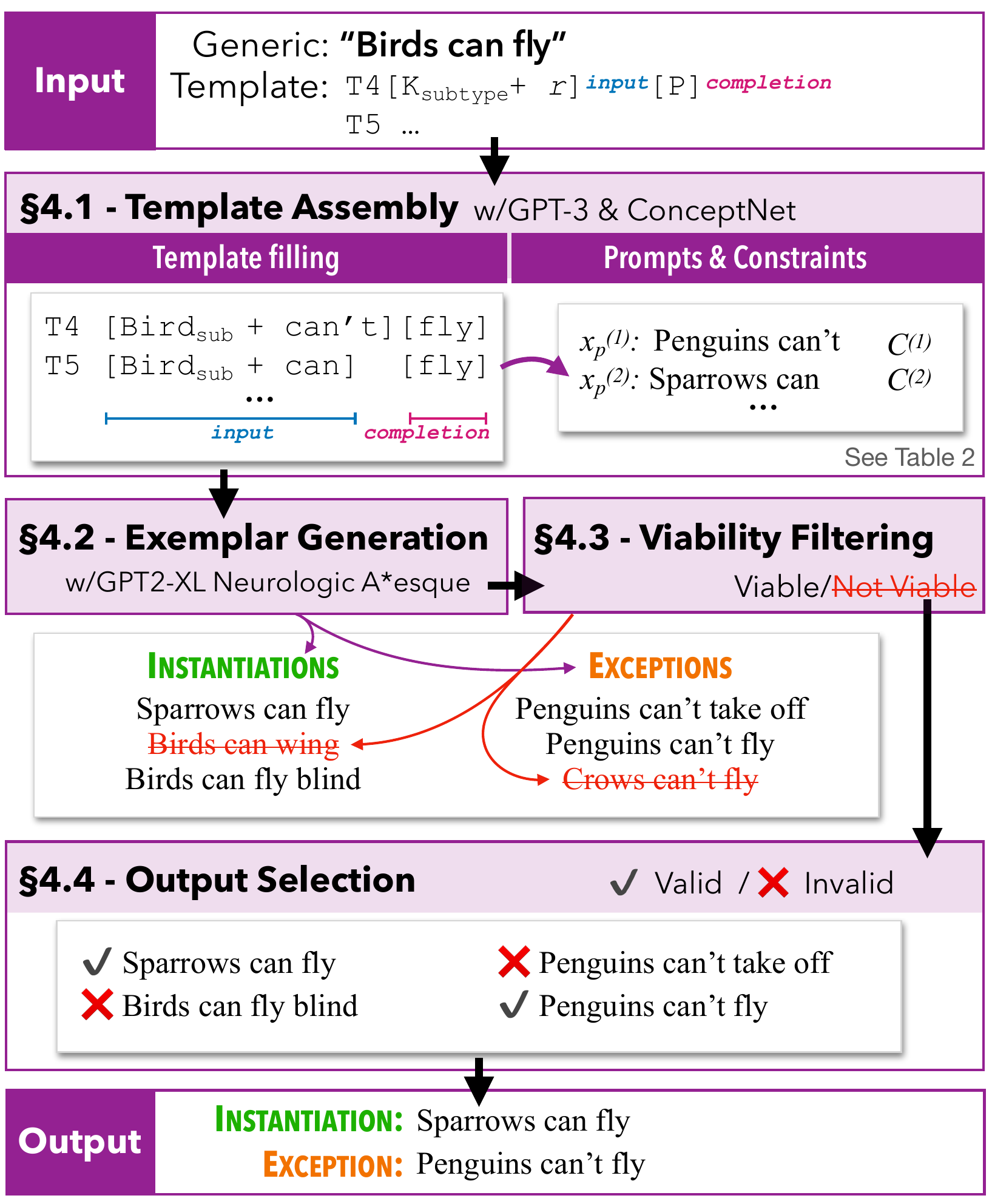}
    \caption{Overview of our method for an input generic.}
    \label{fig:pipeline}
\end{figure}

\section{Methodology}
\label{sec:method}
We propose a pipeline system to automatically generate generics \ev.
Our system takes as input a generic $G$ and
the templates derived from its category\footnote{We assume the generic's category is known.} (\S\ref{sec:tempder}),
and outputs a set of generated \ev (Fig.~\ref{fig:pipeline}). First, the system \textbf{assembles} and populates the templates according to the input generic (\S\ref{sec:tempass}, Fig.~\ref{fig:pipeline}). Then, the filled templates are converted into 
prompts and constraints that control the \textbf{generation} decoding process (\S\ref{sec:gen}). Finally, the output is \textbf{filtered} 
to remove 
non-viable (\S\ref{sec:datafilter}, Fig.~\ref{fig:pipeline}) or 
irrelevant (\S\ref{sec:dataout}, Fig.~\ref{fig:pipeline}) \ev.

\subsection{Template Assembly}
\label{sec:tempass}
To populate our templates (defined in Table~\ref{tab:patts}), we use a dependency parser\footnote{\url{https://spacy.io/}} to identify the text spans of the concept, relation, and property in a generic. Then, \textbf{(i) we extract subtypes} for the concept and property and use these to 
construct the \textbf{(ii)} input i.e., generation prompts $x_p$) and \textbf{(iii)} completion (i.e., lexical constraints $\mathcal{C}$) specifications.

\paragraph{(i) Subtype Extraction}
We extract subtypes using both
ConceptNet~\citep{speer2017conceptnet}\footnote{Relations: \texttt{IsA}, \texttt{InstanceOf}, \texttt{Synonym}}
and GPT-3~\citep{brown2020language}. 
GPT-3 increases the coverage and diversity of subtypes, since 
many natural and valid subtypes may be missing from ConceptNet (e.g., modifier phrases attached to a concept:  ``\textit{young} Arctic fox''). We only use GPT-3 for subtypes of the concept, since by increasing the diversity in the prompt we may encourage diversity in the generated properties (see details Appendix~\ref{app:g3prompts}).

\paragraph{(ii) Input Specification}
We construct the input specifications by constructing generation prompts. Following the template, each prompt consists of either the concept  (or a subtype)  and the relationship (or its negation) (see Table~\ref{tab:patts}). We prepend to each prompt the generic itself and a connective (e.g., ``however''). We rank the prompts by perplexity and 
use the top $k_p$ prompts for generation.

\paragraph{(iii) Completion Specification}
Following the templates, we
constrain the generation output to describe the property (or a subtype) or its \pragneg (see Table~\ref{tab:patts}). We construct a set of completion constraints (e.g., $\mathcal{C}^{(i)}$ in Fig.~\ref{fig:pipeline} specifies ``fly'' should be in the completion) 
using
lexical items including subtypes, synonyms, and morphological forms.

\subsection{Generation}
\label{sec:gen}
In order to generate output that 
has a specific pragmatic relation to the input
without requiring training, we use the NeuroLogic A{\astar}esque (NeuroLogic\astar)~\citep{lu2021neurologic} 
decoding algorithm. NeuroLogic\astar is an unsupervised decoding algorithm that takes as input a prompt $x_p$ and set of lexical constraints $\mathcal{C}$ and produces a 
completion of the prompt
$\hat{y}$ 
which has high likelihood given the prompt \textit{and} high satisfaction of the constraints (estimated throughout the decoding). 
A lexical constraint consists of a set of $n$-grams $w = (w_i^1, \hdots, w_i^m)$ and is satisfied when at least one $w_i \in w$ is in $\hat{y}$ (inclusion constraints) or is not in $\hat{y}$ (exclusion constraints). 

By using
the input prompts (as $x_p$) and completion constraints (as $\mathcal{C}$) derived from our templates (\S\ref{sec:tempass}),
we can control the output content, syntactic form, \textit{and} pragmatic relevance. 
We note that since we cannot concretely define
the set of relevant potential candidates for a property's \pragneg (\S\ref{sec:genlogic}), decoding constraints \textit{must} be used to generate \exceps.

\paragraph{Output Ranking}
We rank the outputs from NeuroLogic\astar by template and prompt and we take the top $k_r$ outputs as potential \ev. 
The outputs are ranked
by perplexity (for fluency) and by the probability of a specific NLI label (for relevance) and we average the two ranks. 
For NLI labels, 
we hypothesize that a good \excep aligns with NLI's contradiction, as does a good \inst with entailment (see Fig.~\ref{fig:pipeline}). While this alignment is useful for ranking, the relationship between the \ev and NLI labels is not this straightforward in reality, as we will discuss (\S\ref{sec:ablation}).

\subsection{Filtering For Viability}
\label{sec:datafilter}
Since pre-trained language models have a tendency to hallucinate facts~\citep{rohrbach-etal-2018-object}
or produce non-specific output (e.g., ``Birds can do things''), we apply a viability filter
to the ranked 
output generations.
Specifically, 
we train a discriminator to predict whether an output is 
viable (i.e., true \textit{and} sufficiently specific that it could be an \evsing) or not,
using human annotated examples (see Appendix~\ref{app:ann} for details). Generations predicted not viable by the trained discriminator are removed from the dataset. 

\subsection{Output Selection}
\label{sec:dataout}
Our final task is to select the generations that are pragmatically 
relevant (i.e., \textbf{valid}; correctly follows a template)
\ev. 
To do this, we first collect gold labels from humans for whether an \ev is 
valid. 
These annotations produce two sets of binary labels; one set each for \insts and \exceps.\footnote{Since a generation that is not an \inst \textit{in not necessarily} an \excep (and vise versa), these cannot be directly combined into a single multi-class labeling task.} 
Although this task is more complex than annotating for viability, removing non-viable generations helps reduce the complexity (i.e., we do not need to worry about false statements that adhere to the template). By annotating only viable generations we also reduce the required amount of annotation.
Using the human annotations, 
we train two 
validity discriminators: one for \exceps, one for \insts. The trained validity discriminators are used to rank and select the best generations for each generic as our system output.

\section{Experiment Details}
\label{sec:datacoll}
We discuss our experimental setup and specify full hyperparameters in Appendix~\ref{app:impdet}.
\subsection{Data Source}
\label{sec:datasrc}
We use a subset of the generics dataset from \citet{Bhagavatula2022I2D2IK}, a set of 30K generics built upon common everyday concepts (e.g., ``hammers'') and relations (e.g., ``used for'') sourced from resources such as GenericsKB \cite{Bhakthavatsalam2020GenericsKBAK} and ConceptNet \cite{speer2017conceptnet}. The dataset includes a diverse variety of concepts, 
including
general knowledge (``Dogs bark''), locative generics (``In a hotel, you will find a bed''), and comparative generics (``Cars are faster than people''). 
We use 653 generics from the test set, 
excluding human referents as the concept (e.g., nationalities, professions) due to social bias concerns.

\subsection{Annotations}
\label{sec:anndetails}
All annotations are done using Amazon Mechanical Turk with three annotators per HIT (paid at $\$15$/hour on average) and processed using MACE~\citep{hovy2013learning} to 
filter annotators
and determine the most likely label. 
We note that while all tasks achieve moderate inter-annotator agreement, the complex pragmatics of generics make these tasks difficult for human annotators.

For \textbf{generic type} (\S\ref{sec:gencat}), we conduct two annotation passes to partition \textit{all 653 generics} into the three groups in Table~\ref{tab:excepttypes}. 
The
Fleiss'~$\kappa$~\citep{fleiss1971measuring} is $0.41$ and $0.58$ for the first and second pass respectively.
Our categorization results in 296 quasi-definitional, 125 principled, and 232 characterizing generics.
For the \textbf{viability filter} (\S\ref{sec:datafilter}), we annotate a set of $7665$ \textit{system generations} from $150$ generics. 
The Fleiss'~$\kappa$ 
is $0.53$.

For \textbf{\evsing gold labels} (\S\ref{sec:dataout}), 
we use separate annotation tasks for \insts and \exceps (see Appendix~\ref{app:ann} for details) with Fleiss' $\kappa$ of $0.40$ and $0.45$ respectively.
For training each discriminator, we randomly sample and annotate ${\sim}1k$ system generations from ${\sim}300$ generics. For human evaluation (\S\ref{sec:humaneval}),  we annotate the \textit{top $5$} discriminator-ranked generations for \textit{all generics} from both our system and the baseline.

\subsection{Discriminators}
For all discriminators, we fine-tune RoBERTa~\citep{liu2019roberta}. All labeled data is split 
such that all generations for a particular generic are in the same data partition. 

\subsection{Few-Shot Baseline}
As a baseline for generation, we use GPT-3~\citep{brown2020language} with few-shot prompting. Since we do not have access to the decoding algorithm for GPT-3, we cannot use decoding constraints to control the output (as in our system). Therefore, we use few-shot prompting in order to control the output of GPT-3.
Specifically, for each template (Table~\ref{tab:patts}) we construct a few-shot prompt (Appendix~\ref{app:g3prompts}) that consists of three examples 
that illustrate the desired template.
This setup is very similar to the prompts to our system, except our system is not provided examples and GPT-3 is not provided with subtypes (when appropriate to the template).
Note, our goal is \textit{not} to produce the best possible generations from GPT-3 but rather to show that constrained generation from GPT-2 (i.e., NeuroLogic\astar) outperforms (and is cheaper and more computationally feasible) than a natural use of GPT-3.

\begin{table}[t]
    \centering
    \scalebox{0.8}{
    \begin{tabular}{ll|rrr}
        \hline
        & & \multicolumn{3}{c}{Subtype Source}\\
        & & G3 & CN & G3+CN\\ \hline
        \multirow{2}{*}{Generated} & Output (\S\ref{sec:gen}) & 42272 & 10496 & 52768\\
        & Viable (\S\ref{sec:datafilter}) & 22865 & 5452 & 28317\\ \hline 

        \multirow{3}{*}{Valid (\S\ref{sec:dataout})} 
        & \Eshort & 4375 & 1922 & 6297\\
        & \Ishort & 10983 & 1579 & 12562\\ \cline{2-5}
        & \textbf{TOTAL} & \textbf{15358} & \textbf{3501} & \textbf{18859}\\

        \hline
    \end{tabular}
    }
    \caption{Statistics of the generated dataset, with GPT-3 (G3) and ConcepNet (CN) subtypes used.}
    \label{tab:datastats}
\end{table}



\begin{table*}[ht]
    \centering
    \scalebox{0.8}{
    \begin{tabular}{l|lll}
    \hline
        & \textbf{Generic} & \textbf{\Inst} & \textbf{\Excep}\\ 
        \hline \hline
        \textbf{(a)} & ``Bleaches may be used to whiten the teeth.'' & \begin{tabular}[t]{@{}l@{}}``non-toxic bleaches can be used\\ to remove discoloration'' \texttt{(t7)} \end{tabular} & \begin{tabular}[t]{@{}l@{}} ``A bottle of liquid bleach should\\ not be used to whiten the teeth'' \texttt{(t4)}\end{tabular}\\
        \hline
        \textbf{(b)} & \begin{tabular}[t]{@{}l@{}} ``A chest pain has a physical cause.''\end{tabular} & \begin{tabular}[t]{@{}l@{}} ``an angina pectoris has an\\ \quad underlying cause'' \texttt{(t5)} \end{tabular} & \begin{tabular}[t]{@{}l@{}} ``a chest pain has an emotional or\\\quad psychological origin'' \texttt{(t1)}\end{tabular} \\ 
        \hline
        \textbf{(c)} & ``A gun are used for hunting.'' & \begin{tabular}[t]{@{}l@{}} ``a shotgun is used for small\\\quad game'' \texttt{(t7)}\end{tabular} & \begin{tabular}[t]{@{}l@{}} ``semiautomatics can be used for\\\quad target practice'' \texttt{(t2)}\end{tabular}\\
        \hline
    \end{tabular}
    }
    \caption{Examples of generated \insts and \exceps. The template used in the prompt for generation is indicated in parentheses (see Table~\ref{tab:patts}).  \vspace{-5pt}}
    \label{tab:dataex}
\end{table*}

\section{Evaluation}
\label{sec:eval}
Using our computational framework,\textbf{ we 
generate 
$18859$ \ev
for $653$ generics} (Table~\ref{tab:datastats}). Example system generations are in Table~\ref{tab:dataex}. 

To evaluate our approach, we first qualitatively investigate our system and outputs (\S\ref{sec:evalqual}) and then conduct a human evaluation (\S\ref{sec:humaneval}). We also conduct a detailed analysis of our system and the implications of our results.(\S\ref{sec:ablation}).
Our results show that our approach produces a large set of high quality generations for this difficult task. They also highlight current limitations in machine reasoning and potential directions for future work.

\subsection{Qualitative Analysis}\label{sec:evalqual}
\paragraph{Observations}
We first observe that while close to half the output generations are untrue or not viable, the majority of viable generations are valid \ev (Table~\ref{tab:datastats}). 
In addition, we see from system outputs (Table~\ref{tab:dataex}) that our system can successfully generate valid \ev with subtypes of both the concept (e.g., ``angina pectoris'' vs. ``a chest pain'' in (b)) and the property (e.g., ``small game'' vs. ``hunting'' in (c)). Furthermore, it produces valid \exceps with both the simpler relation-negation templates (i.e., templates \texttt{t3}/\texttt{t4}; see (a)) \textit{and} with relevant \pragnegs (i.e., templates \texttt{t1}/\texttt{t2}; see (b) and (c)). These highlight the success of our system in producing high-quality \ev.

\paragraph{Discriminator Analysis}
On their respective annotated test sets, the accuracy of the viability 
discriminator (\S\ref{sec:datafilter}) is $75.2$ and the accuracies of the trained validity discriminators are $77.4$ for \insts and $75.0$ for \exceps 

In order to investigate the discriminator quality, we also conduct a manual analysis of the errors made by the \textit{validity} discriminators. We observe that subtypes are particularly difficult for the discriminators to identify (e.g., that ``freshwater lakes and rivers'' are a type of ``water'').
Exoproperties can also be challenging for both the discriminators and humans (e.g., whether ``able to land'' is a subtype or alternative to ``able to move''). 

We also observe that the discriminators identify a number of instances that were mislabeled by the human annotators. In particular, for 10 out of the 22 examples (5 out of the 14) where the \excep (\inst) discriminator prediction disagrees with the human label, we judge the \textit{discriminator} prediction to be correct.
For example, ``clocks are synchronized to the time zone'' is labeled (incorrectly) by humans as an invalid \excep to the generic ``clocks are synchronized to the second'', despite ``to the time zone'' being a \textit{relevant alternative} to ``to the second''. Counting the human-mislabeled instances as correct would increase the discriminator accuracies to 86\% (85\%) for the \exceps (\insts).

\subsection{Human Evaluation}
\label{sec:humaneval}
To quantitatively evaluate our system, we compute precision at $k$ (for $k=1$ and $k=5$) using our human-annotated judgements (\S\ref{sec:anndetails})
(Table~\ref{tab:patk}). 

Our model outperforms the few-shot baseline (i.e., GPT-3) in all cases, and by a large gap (average $12.8$ points). This is especially significant for \exceps, which are more challenging to generate than \insts, and where the baseline performance is close to random. 
Since generics are defaults, it follows that \insts 
should be easier to produce than \exceps. The fact that more generated \insts are true 
($71\%$ versus $40\%$) 
and more true \insts are accepted by the discriminator 
($77\%$ versus $50\%$),
compared to the \exceps, supports this intuition. Hence, the large improvements by our model over the baseline are significant towards generating these difficult \exceps.

Additionally, we examine our model performance across templates.
Specifically, we compute the fraction of generations for a template that annotators label as valid, using the same number\footnote{The models produce similar numbers of generations on all templates except \texttt{t5}.}
of generations for both models for a specific template (Table~\ref{tab:pbypatt}). We see that not only does our model outperform the baseline for the majority of templates, these templates constitute the majority of the generations (`\#Gens' in Table~\ref{tab:pbypatt}). 
\begin{table}[t]
    \centering
    \scalebox{0.9}{
    \begin{tabular}{l|rr|rr}
        \hline
        & \multicolumn{2}{c|}{\Exceps} & \multicolumn{2}{c}{\Insts}\\
        & $P@1$ & $P@5$ & $P@1$ & $P@5$\\
        \hline
        GPT-3 &  0.517 & 0.563 & 0.758 & 0.689\\
        Ours & \textbf{0.632} & \textbf{0.616} & \textbf{0.911} & \textbf{0.882}\\
        \hline
    \end{tabular}
    }
    \caption{Precision at $k$ ($P@k$).}
    \label{tab:patk}
\end{table}

\begin{table}[t]
    \centering
    \scalebox{0.75}{
    \begin{tabular}{l|rrrr|rrr}
        \hline
        & \multicolumn{4}{c|}{\Exceps} & \multicolumn{3}{c}{\Insts}\\
        & \texttt{t1} & \texttt{t2} & \texttt{t3} & \texttt{t4} & \texttt{t5} & \texttt{t6} & \texttt{t7}\\
        \hline
        \#Gens & 401 & 911 & 30 & 43 & 1147 & 4 & 862\\ \hdashline
        GPT-3
        & 0.65 & 0.53 & \textbf{0.52} & \textbf{0.59} & 0.78 & 0.75 & 0.50\\
        Ours & \textbf{0.68} & \textbf{0.54} & 0.30 & 0.47 & \textbf{0.87} & \textbf{1.0} & \textbf{0.87}\\
        \hline
    \end{tabular}
    }
    \caption{Precision by template. \#Gens is per template and is the minimum of the models.     \vspace{-5pt} }
    \label{tab:pbypatt}
\end{table}


The performance comparison by template does not account for the fact that, while our system is constrained to follow the given template, with GPT-3 the template is only suggested by the prompt and so the model output may not adhere to it. As a result, the GPT-3 performance for certain templates (\texttt{t2-4}) is inflated because GPT-3 outputs simpler constructions that do not follow the requested template.
Therefore,  we conduct a manual analysis of the best $40$ baseline (i.e., GPT-3) generations per template, ranked by perplexity. For \exceps, the baseline produces on average only $2.5/40$ generations that fit the desired templates for \texttt{t2}-\texttt{t4}. Additionally, for the one \excep template, \texttt{t1}, where most baseline generations fit the template ($37/40$), our model still outperforms the baseline. For \insts, the baseline performs slightly better (average $10/40$ fitting generations) but still poorly. From this we observe that not only is the baseline not controllable, our model outperforms the baseline in cases when it does adhere to output requirements. 

\subsection{Discussion}
\label{sec:ablation}
\paragraph{Does controllability matter?}
We ablate
the decoding algorithm 
by removing the constraints (i.e., using beam search)
(Table~\ref{tab:beamab}). Although both systems condition their outputs on the same prompts, NeuroLogic\astar, with linguistic-theory-guided constraints, produces over seven times as many unique generations as unconstrained decoding (i.e., beam search).
Additionally, the proportion of valid generations (i.e., accepted by our discriminators) is nearly twice as many for NeuroLogic\astar. This illustrates the importance of 
incorporating linguistic-theory-based control into decoding in order to generate a large set of unique, and valid, \ev.

\vspace{4pt} \noindent \textbf{Do CKBs contain sufficiently rich information?}
We probe whether a CKB (i.e., ConceptNet) contains sufficiently rich type information to produce \ev. Specifically,
we vary the source of subtypes in the template-based prompts and constraints for our system, 
comparing ConceptNet (CN) to 
extracting commonsense knowledge from language models (i.e., from GPT-3 prompting (G3) and GPT-2 masked-language model (MLM)~\citep{devlin2018bert,taylor1953cloze} infilling).

In fact, CN subtypes result in the fewest generations
(Table~\ref{tab:subab}). In contrast, using GPT-3 for subtypes produces the most generations. Although
using MLM for subtypes produces fewer generations than using GPT-3, the proportion of valid generations is comparable and hence MLM could be used as a substitute if using GPT-3 is not feasible. This shows that while CKBs such as ConceptNet are a good source of generics, producing \ev requires knowledge that may not always be encoded within the CKB. Therefore, generating \ev is important for accessing relevant knowledge beyond what is in CKBs and enabling
tools that can effectively use CKBs in reasoning.

\begin{table}[t]
    \centering
    
    \begin{subtable}[t]{1\columnwidth}
        \centering
        \scalebox{0.8}{
        \begin{tabular}{l|ll|ll}
            \hline
            & \multicolumn{2}{c|}{\textbf{Beam}} & \multicolumn{2}{c}{\textbf{NeuroLogic\astar}} \\
            & \#Gens & \%Val & \#Gens & \%Val\\
            \hline
            \textbf{\Eshort} & 5083 & 13.4 & 29962 & 21.0 \\
            \textbf{\Ishort} & 2221 & 39.6 & 22806 & 55.3 \\ \hline
            \textbf{ALL} & 7307 & 21.4 & 52768 & 35.7 \\
            \hline
        \end{tabular}
        }
        \caption{Decoding method ablation: beam search vs. NeuroLogic \astar.}
        \label{tab:beamab}
    \end{subtable}
    \newline
    \vspace*{10pt}
    \newline
    \begin{subtable}[t]{1\columnwidth}
        \centering
        \scalebox{0.77}{
        \begin{tabular}{l|ll|ll|ll}
            \hline
            & \multicolumn{2}{c|}{\textbf{MLM}} & \multicolumn{2}{c|}{\textbf{CN}} & \multicolumn{2}{c}{\textbf{G3}}\\
            & \#Gens & \%Val & \#Gens & \%Val & \#Gens & \%Val\\
            \hline
            \textbf{\Eshort} & 10350 & 25.2 & 7521 & 25.5 & 22441 & 19.5\\
            \textbf{\Ishort} & 4459 & 50.7 & 2975 & 53.0 & 19831 & 55.4\\ \hline
            \textbf{ALL} & 14809 & 32.9 & 10496 & 33.3 & 42272 & 36.3\\
            \hline
        \end{tabular}
        }
        \caption{Subtype ablation: MLM, ConceptNet (CN), GPT-3 (G3).}
        \label{tab:subab}
    \end{subtable}

    \caption{Ablation results. \#Gens: generations after ranking and filtering. \%Val: percent accepted by the corresponding validity discriminator.}
    \label{tab:ablation}
\end{table}

\begin{table}[t]
    \centering
        \scalebox{0.9}{
        \begin{tabular}{l|ll|ll}
            \hline
            & \multicolumn{2}{c|}{\Eshort} & \multicolumn{2}{c}{\Ishort}\\
            & $P@1$ & $P@5$ & $P@1$ & $P@5$ \\
            \hline
            
            Ours & 0.632 & 0.616 & 0.911 & 0.882\\ \hdashline
            + NLI-neu & 0.569 & 0.563 & 0.906 & \textbf{0.891}\\
            + NLI-sim & \textbf{0.839} & \textbf{0.790} & 0.864 & 0.862\\
            + NLI-neu-sim & 0.638 & 0.618 & \textbf{0.913} & \textbf{0.891}\\
            \hline
        \end{tabular}
        }
    \caption{Precision at $k$ with NLI label filtering. NLI-sim is contradiction for \Eshort, entailment for \Ishort }
    \label{tab:nliprec}
\end{table}

            

\vspace{4pt} \noindent \textbf{Does NLI impact \ev?}
Since generics \ev are closely related to default inheritance (nonmonotonic) reasoning, NLI is a natural task for investigating machine reasoning about \ev. Thus, we examine whether controlling the NLI relation between generics and \ev improves precision.
Specifically, we compute the NLI label between the generic (premise) and \evsing (hypothesis) and exclude generations that do not have a specific predicted NLI label: `contradiction' for \exceps and `entailment' for \insts (NLI-sim), `neutral' (NLI-neu), or NLI-sim and `neutral' (NLI-neu-sim). 
We find that by controlling the NLI relation, we improve precision for \exceps by 
20.7
points (Table~\ref{tab:nliprec}). However, for \insts NLI label filtering has a negligible impact on precision.
Therefore, we observe that 
controlling NLI relations can improve \excep quality but is less beneficial for \insts. Additionally, note that the alignment with NLI labels is not actually as straightforward as observed, which we discuss next.

\vspace{2pt} \noindent \textbf{Can NLI sufficiently represent \ev?}
Although we observe an alignment between \textit{predicted} NLI labels and \ev, this actually indicates systematic NLI-model errors, deriving from the insufficiency of NLI schema for 
capturing the nuances of generics \ev. 

\begin{figure}[t]
    \centering
    \includegraphics[width=0.45\textwidth]{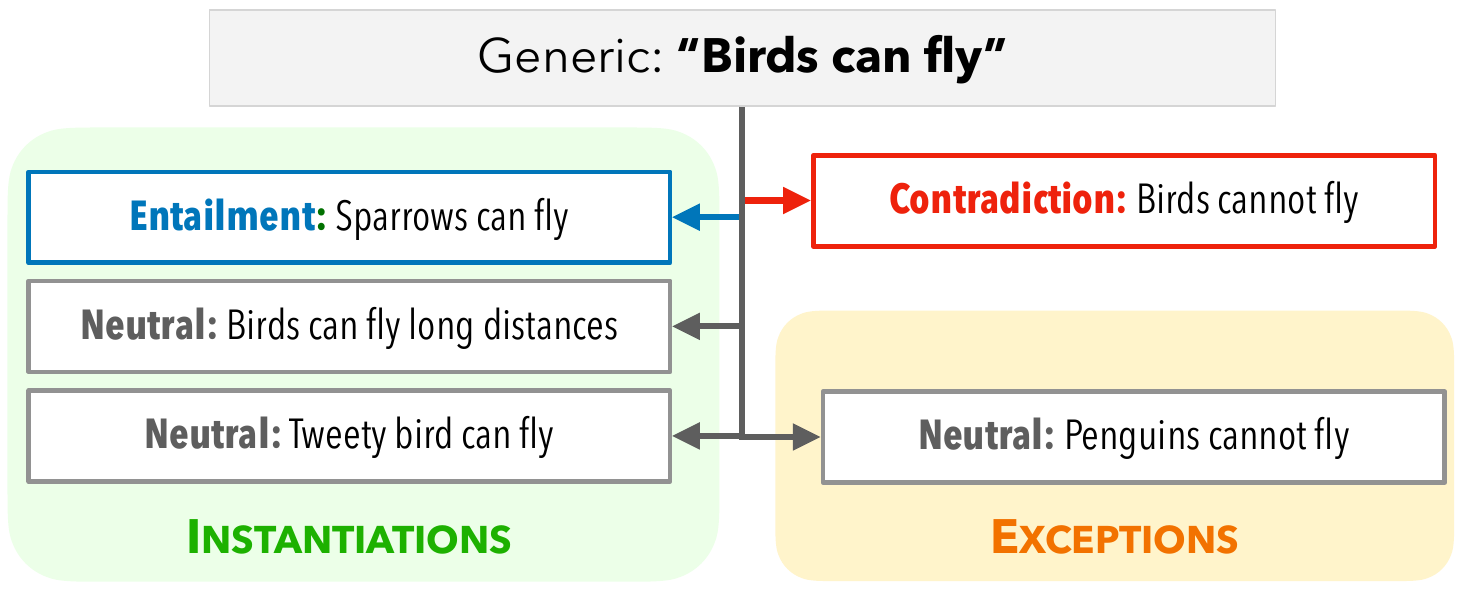}
    \caption{\ev and correct NLI labels.}
    \label{fig:nlifig}
\end{figure}

Consider the sentences
in Fig.~\ref{fig:nlifig}, relating to the generic ``Birds can fly''. We see that \textit{only false} 
statements (i.e., not \exceps) are ``unlikely to be true given the information in the premise [generic]''~\citep{dagan2013recognizing} (i.e., NLI contradictions).
Since the lack of explicit quantification in generics 
does not preclude the existence of exceptions,
\textit{\exceps should actually be labeled neutral} by NLI.
With \insts, we observe that the NLI relationship may be \textit{either} neutral or entailment. 
These theorized alignments, coupled with our prior observations about \ev and predicted NLI labels,
highlight the challenges of reasoning about \ev with NLI.

The examples in Fig.~\ref{fig:nlifig} also highlight that the NLI neutral label does not distinguish between statements that are true but not entailed or contradictory (e.g., ``Penguins cannot fly'') and statements 
with unknown truth value
(e.g., ``Tweety bird can fly''). 
Our
generics \ev
emphasize the 
need for a
more fine-grained notion of NLI.

\section{Conclusion}
\label{sec:conc}
In this work, we draw on insights from linguistics
to propose a novel computational framework to automatically generate valid \ev for generics, as a step towards capturing the nuances of human reasoning for generics.
Our system generates ${\sim}19k$ \ev for 
$653$ generics and 
outperforms 
GPT-3
at generating viable examples, while remaining more controllable. 
We also demonstrate the limitations of CKBs and the importance of explicit linguistic modeling in generating \ev. That is, the importance of linguistic-theory-based decoding and semantics-based filtering with NLI.
Finally, we highlight 
the inability of current NLI models to reason about and represent
the default-inheritance-reasoning relationship between generics and \ev. 

\section*{Acknowledgements}
We would like to thank Sarah-Jane Leslie and Maarten Sap for their helpful discussions, the Beaker Team
at AI2 for
the compute infrastructure, and the anonymous reviewers for their suggestions. This work is supported in part by the National Science Foundation Graduate Research Fellowship under Grant No. DGE-1644869 and by the Office of the Director of National Intelligence (ODNI), Intelligence Advanced Research Projects Activity (IARPA), via the HIATUS Program contract 2022-22072200005. The views and conclusions contained herein are those of the authors and should not be interpreted as necessarily representing the official policies, either expressed or implied, of ODNI, IARPA, the NSF or the U.S. Government. The U.S. Government is authorized to reproduce and distribute reprints for governmental purposes notwithstanding any copyright annotation therein.


\section*{Limitations and Risks}
The generics we source (see \S\ref{sec:datasrc}) are exclusively in English. Therefore, our approach may not be suited to all possible generics in all languages. In particular, our system does not handle generics where valid \insts include negating (\S\ref{sec:genlogic}) the concept. This is due to the restriction that most English generation is left-to-right and it is not possible to define a closed set of possible concept negations for the prompt. 

In this work, we do not generate \ev for generics involving human referents (e.g., professions, nationalities).  We exclude generics involving human referents to mitigate the risk of 
generating socially biased \ev or harmful stereotypes (e.g., ``Black folks go to jail for crimes'' for the generic ``People go to jail for crimes''). Additionally, handling of human stereotypes require methods that are beyond the scope of this paper. For example, a socially-aware \exceps to a generic like ``Girls wear dresses'' would be ``Boys wear dresses, too''. This would require the understanding of the possible subtext of such a statement (e.g. ``Only girls wear dresses''), which is beyond the current capabilities of this study and worthy of future exploration.
    
Finally, we note that while it is not the intended purpose of our system, a malicious user could still use our system to generate \ev for a generic involving a person and propagate potentially harmful social biases.


\bibliography{refs/psychetc,refs/nlp}
\bibliographystyle{formats/acl_natbib}

\appendix

\newpage
\clearpage

\section{Generics}
\label{sec:gens}

\subsection{Generics and Quantifiers}
\label{sec:genquant}

Explicit quantification (e.g., ``\textit{Most} birds can fly'', ``Birds can \textit{usually} fly'') are excluded from this study because the quantifier implicitly accounts for all potential exceptions. That is, by saying ``\textit{Most} birds can fly'' we implicitly indicate that a minority of the birds do not. This being the case, exceptions cannot be generated with statements with explicit quantification. 

\subsection{Generics Definitions}
\label{app:gendef}

\paragraph{Categories of Generics}
We condense the five generic types proposed by \citet{leslie2007generics,leslie2008generics} and \citet{khemlani2009generics} into our three categories (\S\ref{sec:gencat}). The five types are:
\begin{itemize}
\itemsep 0em
    \item \textbf{Quasi-definitional}: generics concerning properties that are assumed to be universal among a concept. This is the same as our quasi-definitional category, see (a) Table~\ref{tab:excepttypes}. The property is considered a defining characteristic of the concept. 
    
    \item \textbf{L-Principled}: generics concerning properties that are prevalent among a concept and are viewed as inherent, or connected in a principled way~\citep{prasada2006principled,prasada2009representation,haward2018development}. These generics are called principled in \citet{leslie2007generics, leslie2008generics}. Note, these generics make up only one half of our ``principled'' category (\S\ref{sec:gencat}). See first example for category (b) in Table~\ref{tab:excepttypes}; the second example there does \textit{not} fit \citet{leslie2007generics,leslie2008generics}'s definition of principled (i.e., L-principled).
    
    \item \textbf{Striking}: generics describing properties that are uncommon and often dangerous, and members of the concept are \textit{disposed} to possess them if given the chance~\citep{leslie2017original}.  For example, the striking generic ``Sharks attack swimmers'' assumes all sharks are capable of attacking swimmers. These generics constitute the second half of our ``principled'' category. See second example (not first) for category (b) in Table~\ref{tab:excepttypes}.
    
    \item \textbf{Majority characteristic}: generics concerning properties that are neither deeply connected to the concept nor striking but occur in the majority of members of the concept. These constitute one half of our ``characterizing category''. See example for (c) in Table~\ref{tab:excepttypes}.
    
    \item \textbf{Minority characteristic}: generics concerning properties that are neither deeply connected to the concept nor striking but occur in the minority of members of the concept. For example, ``Lions have manes'', since only adult male lions (the minority of the lion population) have manes. These constitute the second half of our ``characterizing category''.
\end{itemize}

Both L-principled and striking generics are true in-virtue-of a secondary factor and therefore we group these into one category (i.e., ``principled''; see \S\ref{sec:gencat}). For L-principled generics, this may be a factor that causes the property to occur in the concept (e.g., Birds can fly because they have wings). For striking generics, it is the assumed predisposition of the kind to possess the property if given the chance.

For quasi-definitional generics, because the property is considered defining to concept, there is no implied secondary factor in-virtue-of which the generic is true. Therefore, these generics are descriptive and we put them in a separate category from striking and L-principled generics.

Finally, majority and minority characteristic generics are ambiguous in their interpretation. For example, ``Lions have manes'' can be interpreted as being true in-virtue-of some secondary factor (e.g., as a signal of fitness) or as being a merely accidental relationship. If the interpretation is the former, then lions without manes are valid \exceps (e.g., lion cubs, female lions), while if the interpretation is the latter then then other attributes of lions are valid \exceps (e.g., claws, fur). 

\paragraph{Focuses of Generics}
We note that a generic can focus on the presence of the property within the concept (e.g., ``Birds can fly'' is concerned with which birds can fly) \textbf{or} can focus on the presence of the concept within holders of the property (e.g., ``Triangles have three sides'' is more concerned with what concepts have three sides). We will say that the former kind of generic is \textit{concept-oriented} and the latter is \textit{property-oriented}. A generic can be both concept and property oriented if it is ambiguous between the two readings (e.g., ``Aspirin relieves headaches''). 

In this work, we have discussed and used definitions only for concept-oriented generics.

\subsection{\Ev Definitions}
\label{appsec:evdef}
\paragraph{Interpretations}
As discussed (\S\ref{sec:evlogic}), \exceps counter an interpretation of the generic. Importantly, these interpretations must contain \textit{universal quantification} of either the concept or the property. In this way, the implied universality of the generic can be countered by an \excep. 

We consider the four interpretations of a generic derived from attaching the universal quantifiers ``all'' (or ``always'') and ``only'' to either the concept or the property. For example, for the generic ``Birds can fly'', we have:
\begin{enumerate}
\itemsep 0em
    \item \textbf{\textit{All} birds can fly.}\\$\Rightarrow$ Concept-oriented:\\which birds (i.e., all) can fly.
    \item \textbf{\textit{Only} birds can fly.}\\$\Rightarrow$Property-oriented:\\for being able to fly, how defining (i.e., entirely) are birds.
    \item \textbf{Birds can fly \textit{all ways}.}\\$\Rightarrow$Property-oriented:\\for flight, which types (i.e., all of them) can birds do.
    \item \textbf{Birds can \textit{only} fly.}\\$\Rightarrow$Concept-oriented:\\of the things birds can do, how defining (i.e., entirely) is flying. 
\end{enumerate}
For concept-oriented generics, interpretations 1 and 4 are salient. for property-oriented generics, interpretations 2 and 3 are salient.

The quantifier ``only'' specifies how \textit{defining} the concept is for the property (for concept-oriented generics); for property-oriented generics it specifies how the concept is to the property. Hence, quasi-definitional generics (\S\ref{sec:gencat} correspond to the interpretations containing ``only'' quantifiers. Therefore, their \exceps will counter the implicit assumption that the property is defining for the concept (or vise versa for property-oriented generics). That is, the \exceps will be members of the concept with \textit{other relevant} properties. For example, for the generic ``Stars produce radiation'', an exception is ``The sun produces light''.  

On the other hand, the quantifier ``all'' specifies the prevalence of the property among the concept (for concept-oriented generics). For property-oriented generics, ``all'' specifies the proportion of the property connected with the concept (e.g., how much of the property can members of the concept do). Hence, the ``all'' quantifier corresponds to principled generics (\S\ref{sec:gencat}). Note, that even though striking generics (see \S\ref{app:gendef}) describe very low real-world prevalence, the \textit{implication} is that prevalence is much higher, since individuals are disposed to possess the property~\citep{leslie2017original}.
Therefore, \exceps to principled generics will be members of the concept (or types of the property) that do \textit{not} possess the desired property (or are not present among the concept). For example, a bird that cannot fly (e.g., a penguin) or a type of movement humans cannot do (e.g., fly, for the generic ``humans can move''). 

\paragraph{Logical Forms}
Although we only derived logical forms for concept-oriented generics in this work, 
similar definitions and logical forms can be derived for property-oriented generics.
In particular, only the logical forms for quasi-definitional generics and their \exceps will change if the generic is property-oriented. That is, the $K$ and $P$ in both logicals form for (a) in Table~\ref{tab:excepttypes}) can swapped to obtain the property-oriented versions. In this work, we do not deal with property-oriented generics and their \ev due to the limitations of English generation (i.e., it is left-to-right).

\section{Annotation}
\label{app:ann}
For all annotation tasks, three annotators are used per HIT. When filtering annotators using MACE, we remove annotators with competence below $0.5$ (or the median, if lower).

\begin{figure}[ht]
    \centering
    \includegraphics[width=1\columnwidth]{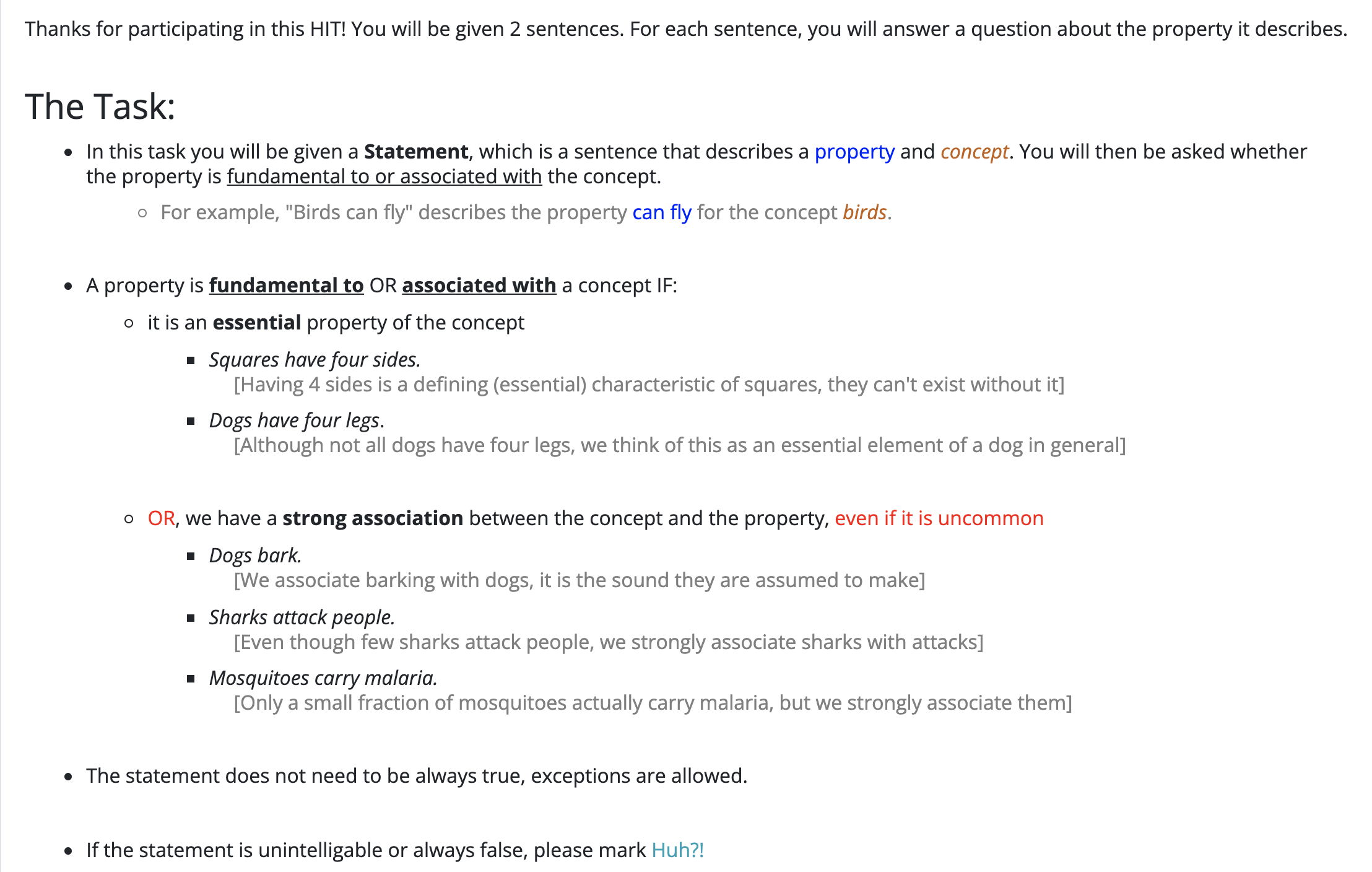}
    \caption{Task instructions for first part of the generic type categorization annotation (\S\ref{sec:anndetails}).}
    \label{appfig:gencat1}
\end{figure}

\begin{figure}[ht]
    \centering
    \includegraphics[width=1\columnwidth]{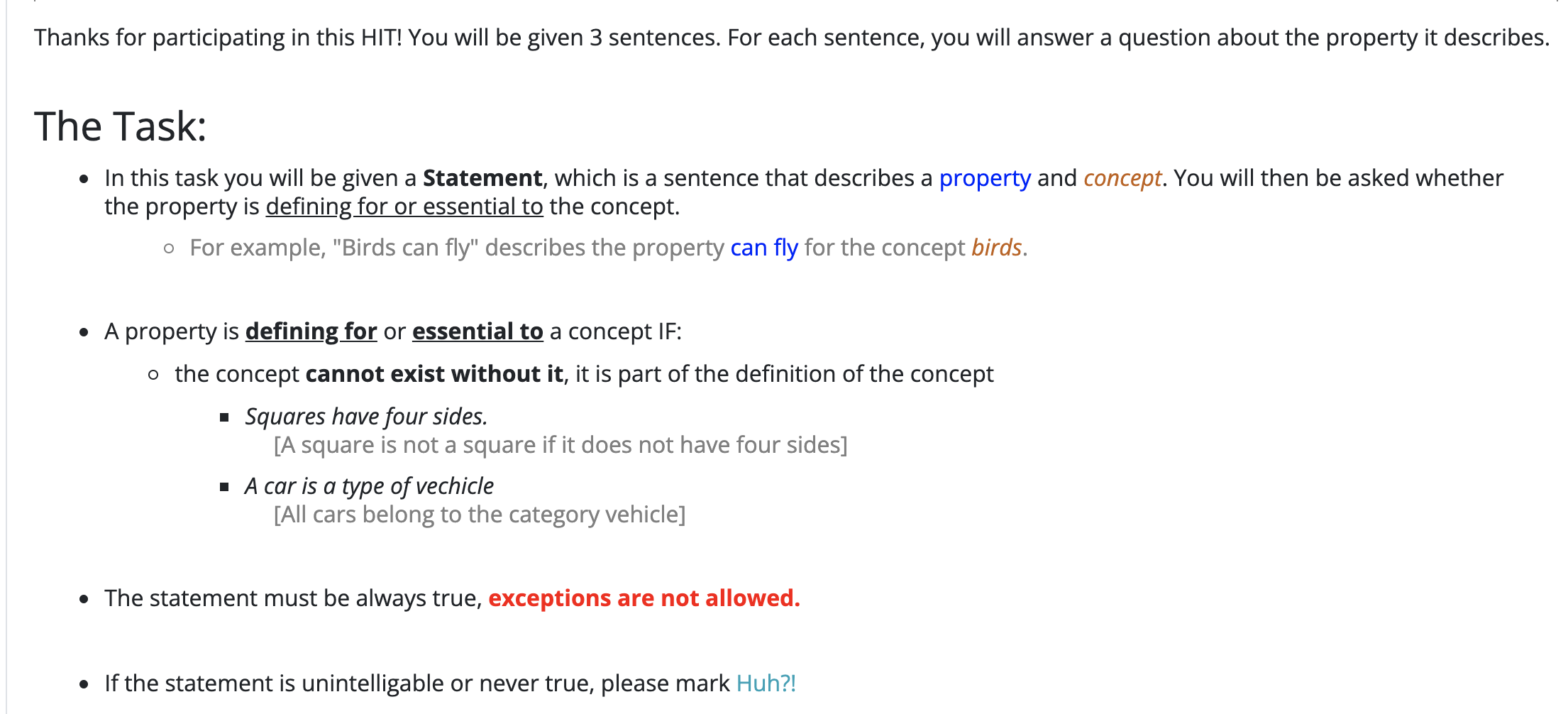}
    \caption{Task instructions for second part of the generic type categorization annotation (\S\ref{sec:anndetails}).}
    \label{appfig:gencat2}
\end{figure}

\paragraph{Generic Type}
Instructions for annotating generic types (\S\ref{sec:gencat}) are shown Figure~\ref{appfig:gencat1} (for the first pass) and Figure~\ref{appfig:gencat2} (for the second pass). The first pass categorizes generics as either characterizing or not (either quasi-definitional or principled). The second pass categorizing non-characterizing generics as either quasi-definitional or principled.

\begin{figure}[ht]
    \centering
    \includegraphics[width=1\columnwidth]{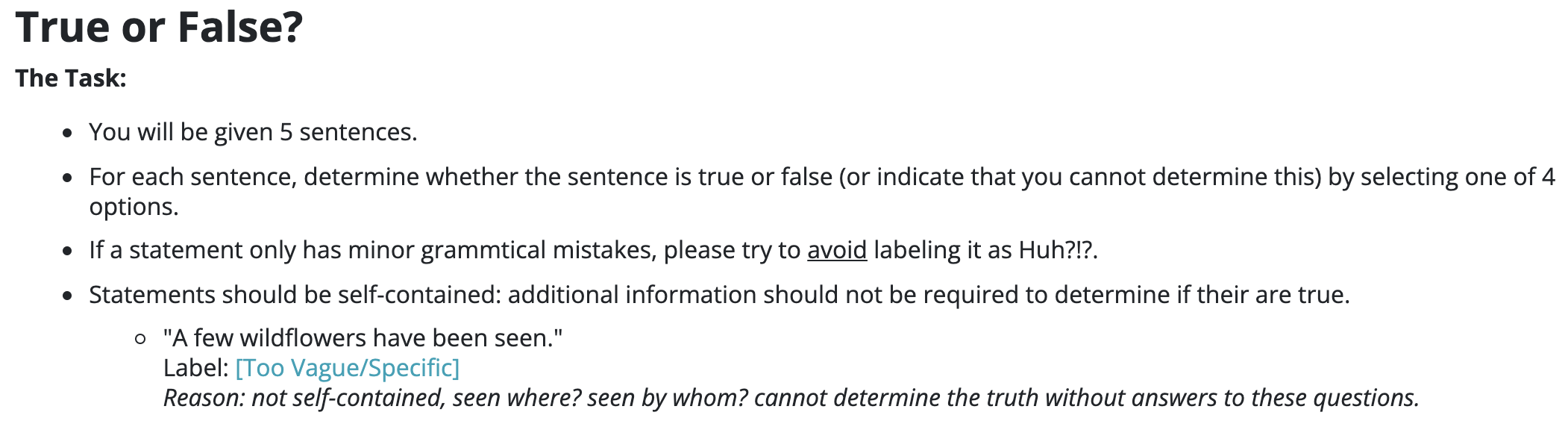}
    \caption{Task instructions for annotating truthfulness (\S\ref{sec:anndetails}).}
    \label{appfig:tftask}
\end{figure}

\paragraph{Viability Task}
Instructions for annotating output generations for viability (\S\ref{sec:datafilter}) are shown in Figure~\ref{appfig:tftask}.

\begin{figure}[ht]
    \centering
    \includegraphics[width=1\columnwidth]{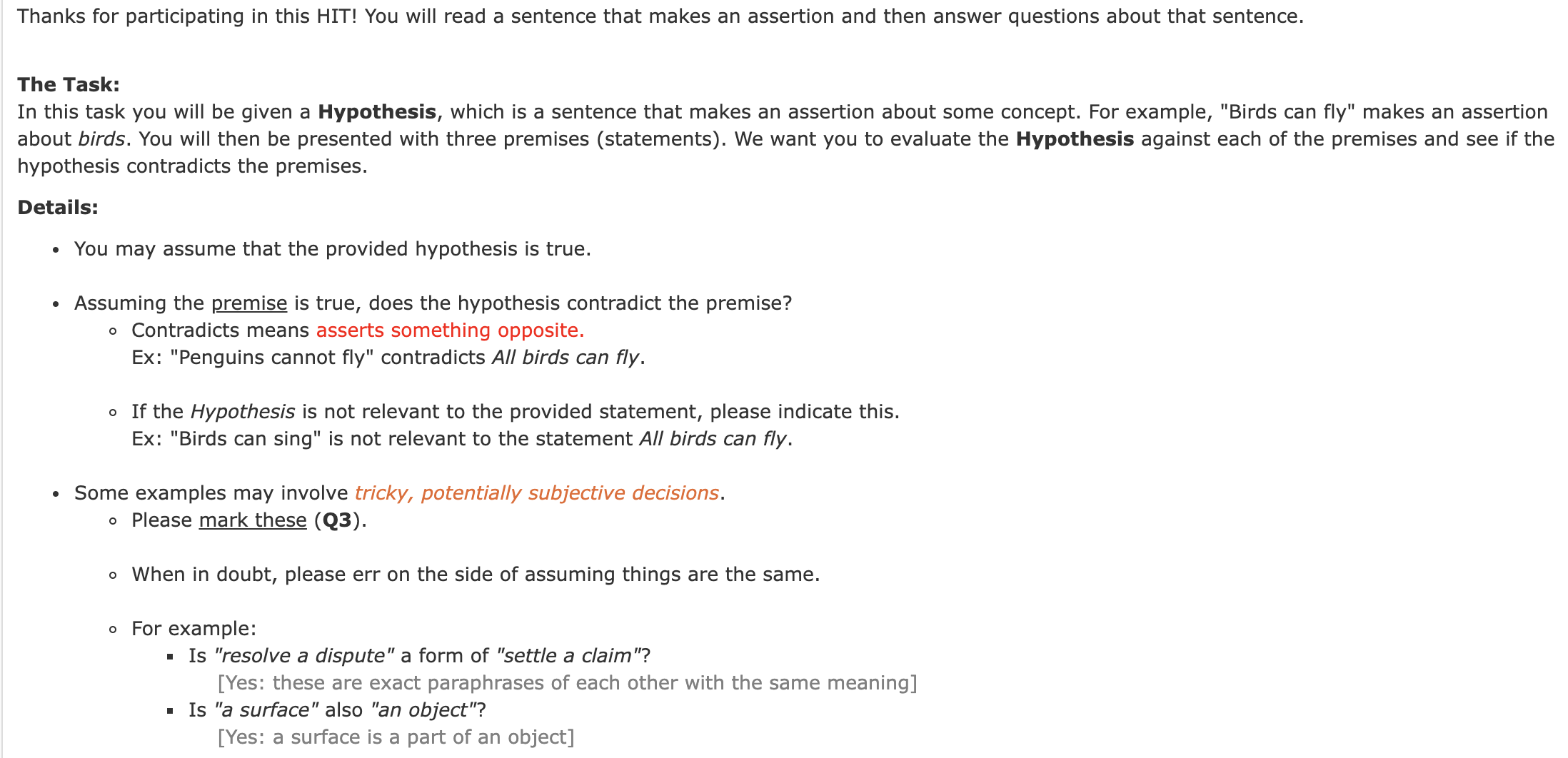}
    \caption{Task instructions for annotating validity of \exceps (\S\ref{sec:anndetails}).}
    \label{appfig:exceps}
\end{figure}

\begin{figure}[ht]
    \centering
    \includegraphics[width=1\columnwidth]{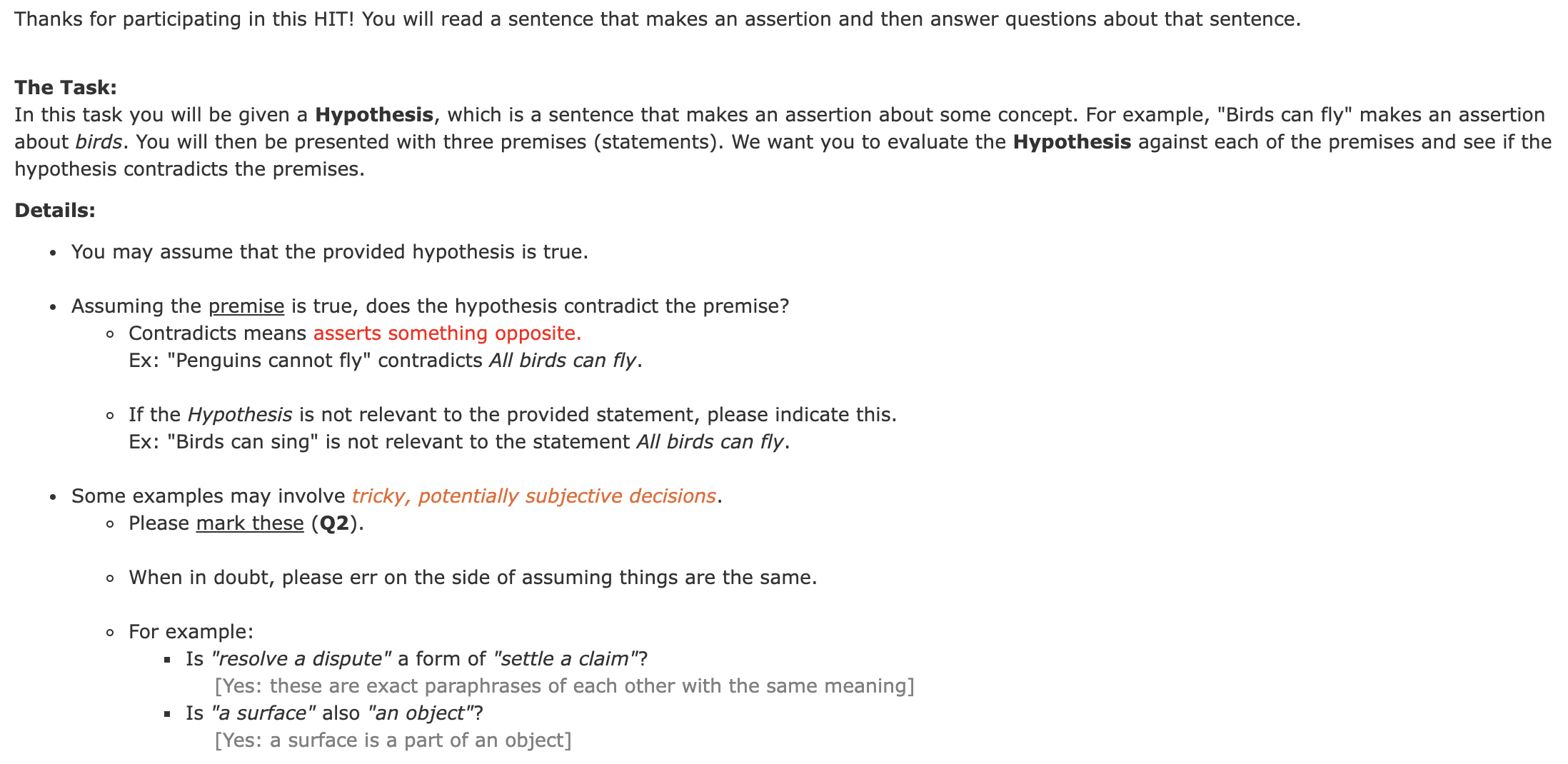}
    \caption{Task instructions for annotating validity of insts (\S\ref{sec:anndetails}).}
    \label{appfig:insts}
\end{figure}

\paragraph{\Ev Gold Labels}
For the \inst template generations, annotators are asked whether the generation contradicts the original generic. Instructions are shown in Figure~\ref{appfig:insts}. However, for the exception template generations, an \excep is not a contradiction of the generic itself but of an associated logical form. For example, ``Penguins cannot fly'' does not actually contradict the generic itself (``Birds can fly'') but a modified form of the generic involving quantification (i.e., ``All birds can fly''). Therefore, we ask annotators whether the generation contradicts two modified forms of the generic. Instructions are shown in Figure~\ref{appfig:exceps}.

We obtain modified forms of the generic by first converting the logical forms in Table~\ref{tab:excepttypes} into a natural language templates by adding a universal quantifier. Then we apply the template to the generic itself. Specifically, from $K(x) \wedge r(x,y) \implies P(y)$ (e.g., for quasi-definitional generics) we derive ``\texttt{[K] [REL] \textit{ONLY} [P]}''. For example, ``mosquitoes drink \textit{only} blood'', which is contradicted by mosquitoes that drink something other than blood. Notice, that exceptions from templates 1 and 2 will contradict these statements. Similarly, for $K(x) \wedge P(y) \implies r(x,y)$ we derive ``\texttt{\textit{ALL} [K] [REL] [P]}''. For example, ``\textit{All} birds can fly'', which is contradicted by birds that cannot fly. Exceptions from templates 3 and 4 will contradict these statements.

\section{Implementation Details}
\label{app:impdet}

\subsection{Data}
We use the generics data from ~\citet{Bhagavatula2022I2D2IK}.
For this study, we source from the subset of the test set found to be valid by the discriminator with probability at least $0.5$ (768 generics). Of these, we exclude all mentions of human referents (e.g., kinship labels, nationalities, titles, professions) and actions (e.g., studying for a test) to arrive at a dataset of 653 generics. We remove human referents using a seed set of human referent terms compiled based on WordNet~\citep{miller1995wordnet} and will be provided with the system code. We remove mentions of actions by excluding generics beginning with ``In order to''. The dataset is licensed under CC-BY and our usage aligns with the intended use of the data. 

\paragraph{Preprocessing}
We remove adverbs of quantification (i.e., usually, typically, generally) from the generics and exclude generics with verbs of consideration (i.e., consider, posit, suppose, suspect, think). We also convert hedging statements to more explicit forms (e.g., ``may have to be'' to ``must be'').

\begin{table}[t]
    \centering
    \begin{tabular}{l|rrr|r}
    \hline
         & Train & Dev & Test & All \\
         \hline
        True & 2831 & 412 & 433 & 3676\\
        False/Non-salient & 3180 & 367 & 442 & 3989\\
        \hline
        Total & 6011 & 779 & 875 & 7665\\
        \hline
    \end{tabular}
    \caption{Data split statistics for truthfulness discriminator (\S\ref{sec:datafilter}).}
    \label{apptab:dstruth}
\end{table}
\begin{table}[ht]
    \centering
    \scalebox{0.85}{
    \begin{tabular}{ll|rrr|r}
        & &  Train & Dev & Test & All\\
         \hline
        \multirow{3}{*}{\small{\excep}} & Valid & 342 & 35 & 35 & 412\\
        & Invalid & 462 & 72 & 53 & 587\\
        \hline
        & Total & 804 & 107 & 88 & 999\\
        \hline\hline
        \multirow{3}{*}{\small{\inst}} & Valid & 374 & 38 & 29 & 441\\
        & Invalid & 466 & 38 & 33 & 537\\
        \hline
        & Total & 840 & 76 & 62 & 978\\
        \hline
    \end{tabular}
    }
    \caption{Data split statistics for validity discriminators (\S\ref{sec:dataout}).}
    \label{apptab:dsev}
\end{table}
\paragraph{Partitions}
The data splits for training the viability discriminator and validity discriminators are shown in Table~\ref{apptab:dstruth} and Table~\ref{apptab:dsev} respectively.

\subsection{Tools}
\label{app:hyps}
For extracting components of the generic data we use spacy\footnote{\url{https://spacy.io/}} for dependency parsing. We use \textit{inflect}\footnote{\url{https://pypi.org/project/inflect/}} to obtain plural and singular word forms and \textit{mlconjug3}\footnote{\url{https://pypi.org/project/mlconjug3/}} to conjugate verbs. We use \textit{nltk}\footnote{\url{https://www.nltk.org/}} for additional synonyms.

\subsection{Hyperparameters}
To obtain subtypes from GPT-3 we use the \textit{davinci} model and top-p sampling with $p=0.9$, temperature $0.8$ and maximum length $100$ tokens. We use the top $5$ sequences to obtain subtypes. For NLI scores, we use RoBERTa fine-tuned on MNLI~\citep{williams2018broad} available from AllenNLP\footnote{\url{https://demo.allennlp.org/textual-entailment/roberta-mnli}}. For the GPT-3 baseline we  use the davinci model and top-p sampling $1.0$, temperature $0.8$, maximum length $50$ tokens and top $5$ sequences. Prompts for GPT-3 are given in Appendix~\ref{app:g3prompts}. GPT2-XL has $1.5$ billion parameters, GPT-3 has $175$ billion parameters. Our experiments are done using Quadro RTX 8000 GPUs.

For generation with NeuroLogic\astar, we use GPT2-XL~\citep{radford2019language} with a maximum length of $50$ tokens and a beam size of $10$ with temperature $10000000$. We set the constraint satisfaction tolerance to $3$. This means that at each step, only candidates whose number of satisfied constraints is within three of the maximum so far are kept. The `look ahead' is also set to $3$; look ahead three generation steps during decoding to estimate future constraint satisfaction. During prompt construction, take the top $k_p = 10$ prompts. If the generic produced less than $10$ prompts total, we take half so that low quality prompts are not used even if few are produced. After ranking the output, we keep the top $k_r = 10$ generations for a template, keeping at most $2$ per prompt. 

\begin{table}[t]
    \centering
    \begin{tabular}{l|r}
        \hline
        \textbf{Parameter} & \textbf{Values}\\
        \hline
        Random seed & 29725\\
        Batch size & [64, 32, 16]\\
        Learning rate & [3e-5, 1e-5, 3e-6]\\
        Number of epochs & [1, 3, 5]\\
        \hline
    \end{tabular}
    \caption{Hyperparameter bounds for the viability discriminator.}
    \label{apptab:hypbounds}
\end{table}
\begin{table}[t]
    \centering
    \scalebox{0.8}{
    \begin{tabular}{l|r}
        \hline
        \textbf{Parameter} & \textbf{Values}\\
        \hline
        Random seed & 4427\\
        Batch size & [64, 32, 16]\\
        Learning rate & [1e-4, 3e-5, 1e-5, 3e-6, 1e-6, 3e-7]\\
        Number of epochs & [1, 3]\\
        \hline
    \end{tabular}
    }
    \caption{Hyperparameter bounds for the validity discriminators.}
    \label{apptab:validhypbounds}
\end{table}
For the viability discriminator, we fine-tune the model for $5$ epochs using a batch size of $16$ and learning rate $1e-5$ and random seed $29725$, selected by manual grid search with $27$ trials (see bounds Table~\ref{apptab:hypbounds}). 

For the validity discriminators, we fine-tune \textit{the viability discriminator} for $3$ epochs with a batch size of $16$ and learning rate $3e-5$. The instantiation discriminator uses a random seed of $4427$ and the exception discriminator $4457$. Hyperparameters are again selected by manual grid search with $36$ trials (see bounds Table~\ref{apptab:validhypbounds}).

\begin{table*}[t]
    \centering
    \begin{tabular}{l|rrr|rrr}
        \hline
        \textbf{Discriminator} & \multicolumn{3}{c}{\textbf{Val}} &  \multicolumn{3}{c}{\textbf{Test}} \\
        & Max & Mean & Var & Max & Mean & Var\\
        \hline
        Viability & 0.771 & 0.685 & 4.3e-3 & 0.752 & 0.673 & 3.6e-3\\
        \hline
        Validity \exceps & 0.757 & 0.600 & 8.8e-3 & 0.750 & 0.557 & 4.4e-3\\
        Validity \insts & 0.763 & 0.582 & 6.6e-3 & 0.774 & 0.577 & 7.1e-3\\
        \hline
    \end{tabular}
    \caption{Trained discriminator accuracy (with mean and variance) on the validation and test sets.}
    \label{apptab:valacc}
\end{table*}
\subsection{Validation Performance}
We show the validation performance for the trained discriminators in Table~\ref{apptab:valacc}.

\section{GPT-3 Prompts}
\label{app:g3prompts}

\subsection{Subtyping}
To obtain subtypes from GPT-3, we first categorize the kinds into six categories: person, animal, other living (e.g., plants), location, temporal (e.g.,  Thursday), and other (e.g., candle, soup) (Table~\ref{apptab:gpt3subs}). For each category, we construct a separate prompt for GPT-3 containing one type and five example subtypes. Then, for each kind we use the prompt from its assigned category to obtain subtypes. Note that we exclude all generics where the kind is ``person''. This is to avoid producing or repeating stereotypes.

To determine the category, we use seed lists, for person, animal, other living, and locative, or the presence of prepositional beginnings (``On'', ``In'', ``At'', ``During''), for locative and temporal. The ``other'' category encompasses all kinds that do not fit into another category.

\subsection{Few-shot Baseline}
The prompts for our few-shot baseline are shown in Table~\ref{apptab:baseprompts}. The three examples in the table are provided each on a separate line. Appended to the prompt is a fourth generic and the necessary connective. The same connective is used across all exception (instantiation) templates and is chosen through manual experimentation. We use ``But also'' for \exceps and ``For example'' for \insts.

\begin{table*}[t]
    \centering
    \begin{tabular}{l|l|l}
        \hline
        Category & Prompt Concept & Prompt Subtypes  \\
        \hline
        Animal & birds &  \begin{tabular}[t]{@{}l@{}}sparrow, canary, large bird,bird of prey, sea bird\end{tabular}\\
        \hline
        Other living & apple tree & \begin{tabular}[t]{@{}l@{}}small apple tree, flowering apple tree,apple tree with ripe apples,\\
                           granny smith apple tree, young apple tree\end{tabular} \\
        \hline
        Locative & hotels  & beach hotel, boutique hotel, resort, bed and breakfast, five star hotel\\
        \hline
        Temporal & day & morning, hot day, short day, afternoon, evening\\
        \hline
        \multirow{2}{*}{Other} & candles & \begin{tabular}[t]{@{}l@{}}scented candle, advent candle, tealight, candle made from beeswax,\\candle that smells floral\end{tabular}\\ \cdashline{2-3}
        & can of soup & \begin{tabular}[t]{@{}l@{}}can of tomato soup, can of mushroom bisque, expired can of soup,\\ unopened can of soup, organic can of soup\end{tabular}\\
        \hline
    \end{tabular}
    \caption{Prompts for generating subtypes with GPT-3.}
    \label{apptab:gpt3subs}
\end{table*}
\begin{table*}[t]
    \centering
    \scalebox{0.8}{
    \begin{tabular}{ll|l}
    \hline
        & Template & Prompt Examples \\ \hline
        (1) & \texttt{[KIND + REL]$^p$ [NEG-PROP]$^{\mathcal{C}}$} & \begin{tabular}[t]{@{}l@{}} Elephants are found in zoos. But also elephants are found in the wild in Africa. \\Viruses are spread through body fluids. But also viruses are spread in the air.\\A hair dryer is used to dry hair. But a hair dryer can also be used to dry clothes.\end{tabular}\\ \hline
        
        (2) & \texttt{[KIND$_{sub}$ + REL]$^p$ [NEG-PROP]$^{\mathcal{C}}$} & \begin{tabular}[t]{@{}l@{}}Elephants are found in zoos. But also African elephants are found in the wild\\\ in Africa.\\Viruses are spread through body fluids. But also coronaviruses are spread\\\ in the air.\\A hair dryer is used to dry hair. But also an electric hair dryer can be used\\ \ to dry clothes.\end{tabular}\\ \hline
        
        (3) & \texttt{[KIND + NEG-REL]$^{p}$  [PROP$_{sub}$]$^{\mathcal{C}}$} & \begin{tabular}[t]{@{}l@{}} Dogs protect buildings from intruders. But also dogs do not protect\\\ apartment buildings from intruders.\\Cowsheds are found on farms. But also cowsheds are not found in orchards.\\The sun produces radiation. But also the sun does not produce x-rays.\end{tabular}\\ \hline
        
        (4) & \texttt{[KIND$_{sub}$ +  NEG-REL]$^p$ [PROP]$^{\mathcal{C}}$} & \begin{tabular}[t]{@{}l@{}} Birds can fly. But also penguins cannot fly.\\Ducks lay eggs. But also male ducks do not lay eggs.\\Dogs protect buildings from intruders. But also very small dogs do not protect\\\ buildings from intruders.\end{tabular}\\ \hline
        \hline
        
        (5) & \texttt{[KIND$_{sub}$ + REL]$^{p}$  [PROP]$^{\mathcal{C}}$} & \begin{tabular}[t]{@{}l@{}}Birds can fly. For example, seagulls can fly.\\Dogs protect buildings from intruders. For example, pitbulls protect buildings\\\ from intruders.\\Ducks lay eggs. For example, female ducks lay eggs.\end{tabular}\\ \hline
        
        (6) & \texttt{[KIND + REL]$^{p}$ [PROP$_{sub}$]$^{\mathcal{C}}$} &\begin{tabular}[t]{@{}l@{}}Viruses are spread through body fluids. For example, viruses are spread\\\  through saliva.\\Dogs protect buildings from intruders. For example, dogs protect some\\\ private homes from intruders.\\Cowsheds are found on farms. For example, cowsheds are found on dairy farms.\end{tabular}\\ \hline
        
        (7) & \texttt{[KIND$_{sub}$ + REL]$^{p}$ [PROP$_{sub}$]$^{\mathcal{C}}$}& \begin{tabular}[t]{@{}l@{}}Birds can fly. For example, Canadian geese fly long distances to migrate.\\Ostriches lay eggs. For example, female ostriches lay large spotted eggs.\\Elephants are found in zoos. For example, African elephants are found in\\\ most large zoos.\end{tabular}\\ \hline
    \end{tabular}
    }
    \caption{Prompts for GPT-3 as Few-shot Baseline. }
    \label{apptab:baseprompts}
\end{table*}


\end{document}